\documentclass[10pt,twocolumn,letterpaper]{article}

\usepackage{cvpr}           

\usepackage{graphicx}
\usepackage{amsmath}
\usepackage{amssymb}
\usepackage{booktabs}
\usepackage{multirow}
\usepackage{makecell} 
\usepackage{tablefootnote}

\def\eg{\emph{e.g.}}

\def\ie{\emph{i.e.}}

\def\etal{\emph{et al.}}

\usepackage[pagebackref=true,breaklinks=true,letterpaper=true,colorlinks,citecolor=blue,urlcolor=blue,linkcolor=blue,bookmarks=false]{hyperref}

\usepackage[capitalize]{cleveref}
\usepackage{float}
\usepackage{blindtext}
\usepackage{graphicx}
\usepackage{caption}
\usepackage{bbding}
\crefname{section}{Sec.}{Secs.}
\Crefname{section}{Section}{Sections}
\Crefname{table}{Table}{Tables}
\crefname{table}{Tab.}{Tabs.}

\def\cvprPaperID{868} 
\def\confName{CVPR}
\def\confYear{2023}

\title{Learning to Dub Movies via Hierarchical Prosody Models}

\begin{document}

\title{Learning to Dub Movies via Hierarchical Prosody Models}

\author{Gaoxiang Cong$^{1}$~~~Liang Li$^{2,3}$\footnotemark[2]~~~Yuankai Qi$^{4}$~~~Zheng-Jun Zha$^{5}$~~~Qi Wu$^{4}$~~~Wenyu Wang$^{1}$\\
~~~Bin Jiang$^{1}$~~~Ming-Hsuan Yang$^{6,7}$~~~Qingming Huang$^{2}$\\
	$^1$Shandong University~~~$^2$Institute of Computing Technology, Chinese Academy of Sciences\\$^3$Lishui Institute of Hangzhou Dianzi University\\~~~$^4$Australian Institute for Machine Learning, University of Adelaide~~~	\\$^5$University of Science and Technology of China~~~$^6$University of California~~~$^7$Yonsei University\\
 }

\maketitle

\begin{abstract}
Given a piece of text, a video clip and a reference audio, the movie dubbing (also known as visual voice clone, V2C) task aims to generate speeches that match the speaker’s emotion presented in the video using the desired speaker voice as reference. 
V2C is more challenging than conventional text-to-speech tasks as it additionally requires the generated speech to exactly match the varying emotions and speaking speed presented in the video. 
Unlike previous works, we propose a novel movie dubbing architecture to tackle these problems via hierarchical prosody modeling, which bridges the visual information to corresponding speech prosody from three aspects: lip, face, and scene. 
Specifically, we align lip movement to the speech duration, and convey facial expression to speech energy and pitch via attention mechanism based on valence and arousal representations inspired by the psychology findings. Moreover, we design an emotion booster to capture the atmosphere from global video scenes. 
All these embeddings are used together to generate mel-spectrogram, which is then converted into speech waves by an existing vocoder.
Extensive experimental results on the V2C and Chem benchmark datasets demonstrate the favourable performance of the proposed method.
The code and trained models will be made available at \href{https://github.com/GalaxyCong/HPMDubbing}{https://github.com/GalaxyCong/HPMDubbing}.

\end{abstract}

\renewcommand{\thefootnote}{\fnsymbol{footnote}}
\footnotetext[2]{Corresponding author.}
\renewcommand{\thefootnote}{\arabic{footnote}}


\section{Introduction}
\label{sec:intro}

\begin{figure}[t]
    \centering
    \includegraphics[width=1.0\linewidth]{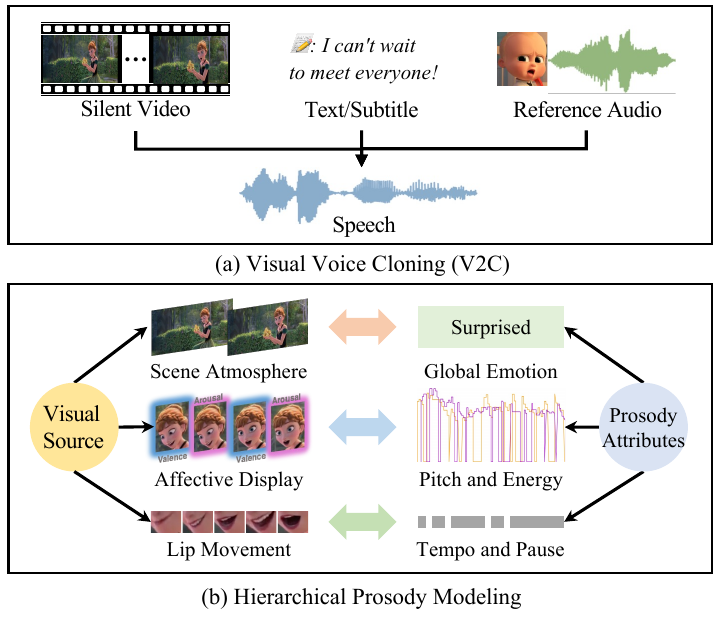}
    \caption{(a) Illustration of the V2C tasks.
    (b) To generate natural speech with proper emotions, we align the phonemes with lip motion, estimate pitch and energy based on facial expression's arousal and valence, and predict global emotion from video scenes.
    }
    \label{fig1}
\end{figure}

Movie dubbing, also known as visual voice clone (V2C)~\cite{chen2022v2c}, aims to convert a paragraph of text to a speech with both desired voice specified by reference audio and desired emotion and speed presented in the reference video as shown in the top panel of Figure~\ref{fig1}.
V2C is more challenging than other speech synthesis tasks in two aspects: 
{first}, it requires synchronization between lip motion and generated speech;
{second}, it requires proper prosodic variations of the generated speech to reflect the speaker's emotion in the video (\ie, the movie's plot).
These pose significant challenges to existing voice cloning methods.

Although significant progress has been made, existing methods do not handle the challenges in V2C well. 
Specifically, text-based dubbing methods~\cite{ren2020fastspeech, shen2018natural, wang2017tacotron, ren2019fastspeech}  construct speeches from given text conditioned on the different speaker embedding but do not consider audio-visual synchronization. 
On the other hand, lip-referred dubbing schemes~\cite{ hegde2022lip, wang2022fastlts, lu2022visualtts}  predict mel-spectrograms directly from a sequence of lip movements typically by encoder-decoder models. 
Due to high error rates in generated words, these methods can hardly guarantee high-quality results. 
Furthermore, video-and-text based dubbing methods~\cite{hu2021neural, hassid2022more, lu2022visualtts} focus on inferring speaker characters (\eg, age and gender). 
However, these visual references usually do not convey targeted emotion well as intended in V2C.

An ideal dub should align well with the target character so that the audiences feel it is the character speaking instead of the dubber~\cite{chaume2007dubbing}. 
Thus, a professional dubber usually has a keen sense of observing the unique characteristics of the subject and acts on voice accordingly. 
In this work, we address these issues with a hierarchical dubbing architecture to synthesize speech. 
Unlike previous methods, our model connects video representations to speech counterparts at three levels: lip, face, and scene, as shown in Figure~\ref{fig1}.

In this paper, we propose a hierarchical prosody modeling for movie dubbing, which could keep the audio-visual sync and synthesis speech with proper prosody following the movie's plot.
Specifically, we first design a duration alignment module that controls speech speed by learning temporal correspondence via multi-head attention over phonemes and lip motion.
Second, we propose an affective-display based Prosody Adaptor (PA), which learns affective psychology computing conditioned on facial expression and is supervised by corresponding energy and pitch in the target voice. 
In particular, we introduce arousal and valence features extracted from facial regions as emotion representations. 
This is inspired by the affective computing method~\cite{toisoul2021estimation}, which analyses the facial affect relying on dimensional measures, namely valence (how positive the emotional display is) and arousal (how calming or exciting the expression looks).
Third, we exploit a scene-atmosphere based emotion booster, which fuses the global video representation with the above adapted hidden sequence and is supervised by the emotive state of the whole voice. 
The outputs of these three modules are fed into a transformer-based decoder, which converts the speech-related representations into mel-spectrogram.
Finally, we output the target speech waves from the mel-spectrogram via a powerful vocoder.

The contributions of this paper are summarized below:
\begin{itemize}
\vspace{-2mm}
    \item We propose a novel hierarchical movie dubbing architecture to better synthesize speech with proper prosody by associating them with visual counterparts: lips, facial expressions, and surrounding scenes. %
    \vspace{-2mm}
    \item We design an affective display-based prosody adaptor to predict the energy and pitch of speech from the arousal and valence fluctuations of facial regions in videos, which provides a fine-grained alignment with speakers' emotions. %
    \vspace{-2mm}
    \item Extensive experimental results demonstrate the proposed method performs well against state-of-the-art models on two benchmark datasets. %
\end{itemize}

\section{Related Work}
\noindent\textbf{Text to Speech Synthesis}. 
Over the recent years, numerous TTS models ~\cite{ping2018clarinet, ping2017deep, shen2018natural, wang2017tacotron, arik2017deep, li2019neural, ren2019fastspeech} have been proposed for generating high-quality natural speech conditioned on given text. 
Tacotron~\cite{wang2017tacotron} is an end-to-end generative TTS model that synthesizes speech directly from characters. 
Then, Tacotron2~\cite{li2019neural} replaces the RNN structures by introducing the attention mechanism to improve training efficiency and solve the long dependency issue. 
Furthermore, FastSpeech~\cite{ren2019fastspeech} and Fastspeech2~\cite{ren2020fastspeech} exploit the Feed-Forward Transformer (FFT) to generate mel-spectrogram from phoneme sequences. 
Despite the impressive voice generated, these methods cannot provide the audio with desired emotion and audio-visual sync for movie dubbing.

\smallskip \noindent\textbf{Lip to Speech Synthesis.} 
This task aims to reconstruct speech based on the lip motions alone~\cite{kumar2019lipper, assael2016lipnet}.
Lip2Wav~\cite{prajwal2020learning} is a sequence-to-sequence architecture focusing on learning mappings between lip and speech for individual speakers. 
Recently, \cite{hegde2022lip, wang2022fastlts, mira2022svts, shi2022learning} improve the architecture and training methods, and provide the possibility of unconstrained speech synthesis in the wild. 
However, lip-to-speech is incompetent for movie dubbing because the word error rate is still high~\cite{afouras2018deep, assael2016lipnet, son2017lip, chung2017lip, ephrat2017vid2speech}. 
In this work, we focus on reconstructing accurate speech from lip motions and generating the desired emotion and identity with proper prosody.

\smallskip \noindent\textbf{Talking Heads.} 
Numerous methods have been developed for audio-visual translation~\cite{yang2020large} or speaking style transfer~\cite{xie2021towards} by reconstructing the visual content in video~\cite{zhou2019talking, chen2019hierarchical,zhou2020makelttalk,liang2022expressive,song2022everybody, wang2022one, yin2022styleheat, liu2022learning, zhou2022audio}. 
Wav2Lip~\cite{prajwal2020lip} uses an expert lip-syncs discriminator to morph lip movements of arbitrary identities.
Recently, Papantoniou~\etal~\cite{papantoniou2022neural} develop a Neural Emotion Director (NED) to manipulate emotions while preserving speech-related lip movements. 
However, these methods cannot adapt to the movie dubbing task because they emphasize using generative models to readjust the facial regions instead of reconstructing the desired speech. 

\begin{figure*}[!htbp]
    \centering
    \includegraphics[width=1.0\linewidth]{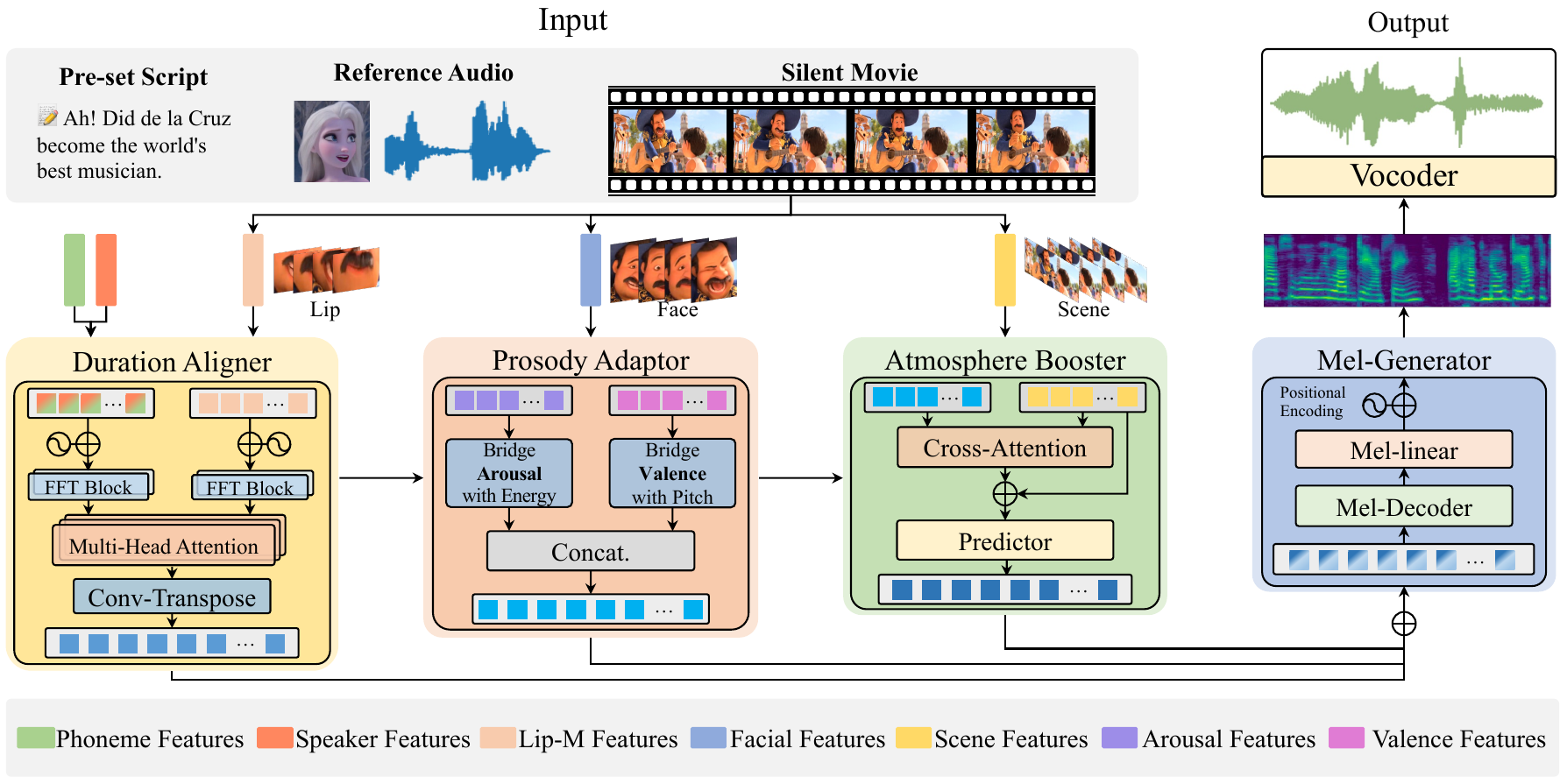}
    \caption{Architecture of the proposed hierarchical modular network for movie dubbing, which consists of four main components: Duration Aligner (Sec.~\ref{sec:da}), which learns to predict speech duration based on aligning lip movement and text phoneme; Prosody Adaptor (Sec.~\ref{sec:pa}), which predicts energy and pitch from facial arousal and valence, respectively; Atmosphere Booster (Sec.~\ref{sec:sb}), which learns a global emotion embedding from a video scene level; and Mel-Generator (Sec.~\ref{sec:ag}), which generates mel-spectrograms from embeddings obtained by the aforementioned three modules. The mel-spectrograms are finally converted to audio by a widely adopted vocoder.
    }
    \vspace{-8pt}
    \label{fig2}
\end{figure*}

\smallskip \noindent\textbf{Visual Voice Cloning.}
Movie dubbing, also known as visual voice clone, aims to convert scripts to speech with both desired voice identity and emotion based on the reference audio and video. 
To control the speed of generated speech, Neural Dubber~\cite{hu2021neural} exploits a text-video aligner by using scaled dot-product attention mechanism. 
VTTS~\cite{hassid2022more} uses multi-source attention to fuse the triplets feature and outputs the mel-spectrogram via an RNN-based decoder. 
Since explicit emotion categories~\cite{li2019hierarchical} do not exist in these methods, Chen~\etal~\cite{chen2022v2c} develops a V2C model on a more challenging Densiny Animation dataset, which concentrates on emotional dubbing for movie characters. 
Although the V2C considers emotion labels, the adopted global video representation negatively affects the fine-grained emotional expression and makes it challenging to render correct prosody corresponding to plot developments. 
To solve this issue, we propose a hierarchical movie dubbing architecture to better synthesize speech with proper prosody and emotion.

\section{Method}
The main architecture of the proposed  model is shown in Fig.~\ref{fig2}.   
First, we use a phoneme encoder~\cite{chen2022v2c} to convert the input text $Z_{text}$ to a series of phoneme embeddings $\mathcal{O}=\{\mathbf{o}_1, ..., \mathbf{o}_L\}$ and use a speaker encoder $F_{spk}$~\cite{chen2022v2c} to capture the voice characteristics $\mathcal{U}$ from different speakers. 
Then, taking phonemes and lip regions as input, the duration aligner module uses a multi-head attention mechanism to learn to associate phonemes with related lip movements. 
Next, the affective display-based prosody adaptor module learns to predict the energy and pitch of the desired speech based on arousal and valence features extracted from facial expressions, respectively.  %
And then, the scene atmosphere booster encodes a global representation of emotion of the entire video content.
All the outputs of the above three modules are combined to generate mel-spectrograms, which are finally transformed to a waveform $Y_{voice}$ using a adopted vocoder. We detail each module below.

\subsection{Duration Aligner}\label{sec:da}
The duration aligner contains three steps: (1) extracting the lip features from movie; (2) aligning the phonemes of text with the lips; (3) expanding the fused phoneme-lip representation to the desired mel-spectrogram length.

\smallskip \noindent\textbf{Extracting lip feature.} Let $D_w$, $D_h$ and $D_c$ be the width, height and number of channels of the video frames, respectively. We first extract lip regions $\mathbf{x_m}\in \mathbb{R}^{{T_v}\times{D_w}\times{D_h}\times{D_c}}$ from the given video using mouth region pre-processing from~\cite{hu2021neural, ma2020towards, martinez2020lipreading, ma2021lip, ma2022training}. 
Then we exploit the LipEncoder
to obtain the lip movement representation:
\begin{equation}
    \begin{aligned}
        \mathbf{E_{lip}}= \mathrm{LipEncoder}(\mathbf{x_m}) \in \mathbb{R}^{{T_v}\times{D_m}},
     \end{aligned}
\end{equation}
where $T_v$ denotes the number of video frames, and $D_m$ is the hidden dimension of the dynamic lip feature. 
The LipEncoder consists of several feed-forward transformer blocks that are suitable for capturing both long-term and short-term dynamics lip movement features.

\smallskip \noindent\textbf{Aligning  text with lips}. 
Inspired by the success of  attention mechanism for cross-modality alignment~\cite{hop+,recurrent,liangli_long, cong_lsgan,diagnose,rgan,compress, Liang_Distributed},
we adopt multi-head attention to learn the alignment between the text phoneme and the lip movement sequence. 
We use lip embedding as a query to compute the attention on text phonemes. The larger the attention, the more related between a lip embedding and a text phoneme. 
Due to variations of mouth shapes and pronunciations, the multi-head attention mechanism is suitable for learning their alignments from different aspects. 
The text-video context sequence $\mathbf{E_{{lip,txt}}}$ = $[\alpha_{{lip,txt}}^1,... ,\alpha_{{lip,txt}}^n]\in \mathbb{R}^{{T_v}\times{D_m}}$ is a concatenation of outputs of $n$ attention heads. Concretely, the $k$-th head's output $\alpha_{{lip,txt}}^k$ is obtained by:
\begin{equation}
\small{
\begin{split}
\alpha_{{lip,txt}}^k={\mathrm{softmax}}(\frac{\mathrm{Q}^{\top}{\mathrm{K}}}{\sqrt{d_{k}}}+\mathrm{M_t})\mathrm{V}^{\top},\\ \mathrm{Q}=\mathrm{W}{_j^Q}{E_{lip}}^{\top},\mathrm{K}=\mathrm{W}{_j^K}{\mathcal{O}}^{\top},\mathrm{V}=\mathrm{W}{_j^V}{\mathcal{O}}^{\top},
\end{split}
}
\end{equation}
where $\mathbf{W}{_j^*}$ is a learnable parameter matrix, $d_{k}$ is the embeeding dimension of $\alpha_{{lip,txt}}^k$, and $\mathbf{M_t}$ is a mask matrix indicating whether a token can be attended. 
The aligned representation $E_{lip,txt}$ is later expanded to the length of the desired mel-spectrogram.

\smallskip \noindent\textbf{Expanding to the desired length.} %
According to the findings in~\cite{hu2021neural},
in an audio-visual clip, the length of a mel-spectrograms sequence is $n$ times that of a video frame sequence because they are temporally synchronized in audio and visual modalities. The number $n$ is computed as:
\begin{equation}
    \begin{aligned}
        n = \frac{T_{mel}}{T_v}=\frac{sr/hs}{FPS} \in \mathbb{N}^{+},
     \end{aligned}
\end{equation}
where $FPS$ denotes the Frames per Second of the video, $sr$ denotes the sampling rate of the audio, and $hs$ denotes hop size  when transforming the raw waveform into mel-spectrograms. 
Phoneme-lip feature $E_{lip,txt}$ is simply duplicated $n$ times in~\cite{hu2021neural} as the final phoneme-lip representation, which lacks flexibility.
Instead, we propose to use transposed convolutions to learn the expansion of $E_{lip,txt}$, 
which can be formulated as:
\begin{equation}
\small{
    \mathbf{M}_{pho,lip} = \mathrm{Conv\text{-}Transpose}(n,E_{{lip,txt}})\in \mathbb{R}^{{T_y}\times{D_m}},
    }
\end{equation}
where $T_y$ denotes the length of the desired mel-spectrogram, $D_m$ is the dimension of the initial mel-spectrogram. The parameters (stride and kernel size) of the transposed convolution can be set under the guidance of $n$ so that $T_y\approx n\times T_v$.

\subsection{Affective-display based Prosody Adaptor}\label{sec:pa}
One of the critical issues in the V2C task is to describe the speaker's emotions in the given video. 
To solve this problem, we design an affective-display based Prosody Adaptor (PA), which uses the arousal and valence extracted from facial expressions to represent the emotion. The arousal and valence are then used to predict the energy and pitch of the desired speech model. 

\smallskip \noindent\textbf{Valence and Arousal Feature.} 
To accurately capture the valence and arousal information from facial expressions, we utilize an emotion face-alignment network (EmoFAN)~\cite{toisoul2021estimation} to encode the facial region into valence $\textbf{V}$ and arousal $\textbf{A}$.
$
    \begin{aligned}
        \textbf{V}, \textbf{A} = \mathrm{EmoFAN}(x_f) \in \mathbb{R}^{{T_v}\times{D_m}},
     \end{aligned}
$
$x_f\in \mathbb{R}^{{T_v}\times{D_w}\times{D_h}\times{D_c}}$ is the face region extracted via $S^{3}FD$ face detection~\cite{zhang2017s3fd}. The EmoFAN focuses on facial regions relevant to emotion estimation, which utilizes a face alignment network (FAN) for facial point detection to ensure robustness by jointly predicting categorical and continuous emotions.

\noindent\textbf{Bridging Arousal with Energy}. 
Arousal is the physiological and psychological state of being awoken or of sense organs stimulated to the point of perception. 
To bridge the vocal energy  with arousal display, we compute an arousal context vector $A_i^l$ for frame-level energy of the desired speech from phoneme-lip representation $\mathbf{M}_{pho,lip}$:
\begin{equation}
\begin{split}
{A}_i^l&={\sum \limits_{k=0}^{{T_y}-1}\xi_{i,k}{{M}^k_{pho,lip}}},\\ 
\xi_{i,k}&={{\exp (\hat{\xi}_{i,k})}/{\sum \limits_{j=0}^{T_y-1} \exp (\hat{\xi}_{i,j})}},\\
\hat{\xi}_{i,k} &= \mathbf{w^{\top}_a}\tanh(\mathbf{W^{\top}_a}{A}_i+\mathbf{U^{\top}_a}{{M}^k_{pho,lip}}+\mathbf{b_a})
\end{split}
\end{equation}
where $i$ is frame index, $A_i$ is the $i$-th row of $\mathbf{A}$, $\xi_{i,k}$ is the attention weight on the $k$-th phoneme-lip feature $M_{pho,lip}^k$ (the $k$-th row of $\mathbf{M}_{pho,lip}$) regarding $i$-th arousal display; 
$\mathbf{w^T_a}$, $\mathbf{W^T_a}$, $\mathbf{U^T_a}$ and $\mathbf{b_a}$ are learnable parameters; $T_y$ is the length of the desired mel-spectrogram.  

Then, we use an energy predictor to project the arousal-related phoneme-lip embedding ${A}_i^l$ to the energy of speech:
\begin{equation}
    \begin{aligned}
        {E_{aro}} = \mathrm{Predictor}(\{{A}_i^l\}_{i=1}^{T_y}),
     \end{aligned}
\end{equation}
where ${E_{aro}}\in \mathbb{R}^{{T_y}}$ represents the predicted energy of speech, and the energy predictor consists of several fully-connected layers,  Conv1D blocks and layer normalization.

\smallskip \noindent\textbf{Bridging  Valence with Pitch}. 
In prosody linguistics, speakers can speak with a wide pitch range (this is usually associated with excitement) while, at other times, with a narrow range. %
Valence is the affective quality referring to the intrinsic positiveness or negativeness of an event or situation. %
Similar to arousal to energy, we first compute a valence context vector for the frame-level pitch of the desired speech from phoneme-lip representation $\mathbf{M}_{pho,lip}$:
\begin{equation}
\begin{split}
{V}_i^l&={\sum \limits_{k=0}^{{T_y}-1}\psi_{i,k}{M}^k_{pho,lip}},\\ 
\psi_{i,k}&={{\exp (\hat{\psi}_{i,k})}/{\sum \limits_{j=0}^{{T_y}-1} \exp (\hat{\psi}_{i,j})}},\\
\hat{\psi}_{i,k} &= \mathbf{w^{\top}_g}\tanh (\mathbf{W^{\top}_g}{V}_i+\mathbf{U^{\top}_g}{M}^k_{pho,lip}+\mathbf{b_g})
\end{split}
\end{equation}
where $\mathbf{w^T_g}$, $\mathbf{W^T_g}$, $\mathbf{U^T_g}$ and $\mathbf{b_g}$ are learnable parameters. Then, we exploit a pitch predictor to convert the valence-related phoneme-lip embedding ${V}_i^l$  to the pitch of speech, which can be formulated as:
\begin{equation}
    \begin{aligned}
        {P_{val}} = \mathrm{Predictor}(\{{V}_i^l\}_{i=1}^{T_y}).
     \end{aligned}
\end{equation}

Finally, we concatenate the two prosody-related features as the contextualized affect primitives:
\begin{equation}
    \begin{aligned}
        \mathbf{M_{p}}=  
        \{{[{A}_i^l;{V}_i^l]}\}_{i=1}^{T_y}.
     \end{aligned}
\end{equation}

\subsection{Scene Atmosphere Booster}\label{sec:sb}
As a unique form of artistic expression, the scene layout and colours of the film convey an emotional atmosphere to evoke resonance with the audience~\cite{wehrmann2017movie, burghardt2016beyond}. 
To reason the comprehensive emotion, we design a scene atmosphere booster to combine the global context information and generated prosody.
First, we use the I3D model~\cite{carreira2017quo} to extract the scene representation $\mathcal{S}$ from the video.
Then, we fuse the global contextual emotion of vision with prosody information of speech through a cross-model attention mechanism by:
\begin{equation}
    \mathbf{M_e}=\mathrm{Softmax}(\frac{{\mathbf{M_{p}}}{S^\top}}{\sqrt{D_m}})\mathbf{M_{p}}\in \mathbb{R}^{{T_y}\times{D_w}}.
\end{equation}

Finally, formulated as a maxpool and fully-connected layer, the emotional predictor projects prosody-scene context sequence $\mathbf{M_e}$ to emotional embedding:
\begin{equation}
    {G_{emo}}=\mathrm{Maxpool}(\mathrm{Predictor}(\mathbf{M_e})).
\end{equation}

\subsection{Audio Generation}\label{sec:ag}
We fuse the three kinds of speech attribute information by concatenation operation in the spatial dimension. 
Then, we use transformer-based mel-spectrogram decoder to convert the adapted hidden sequence into mel-spectrogram sequence in parallel by:
\begin{equation}
    \begin{aligned}
        \mathbf{f} = \mathrm{TransformerDecoder}(\mathbf{M}_{pho,lip}\oplus{\mathbf{M_{p}}}\oplus{\mathbf{M_{e}}}),
     \end{aligned}
\end{equation}
where $M_{pho,lip}$, $M_{p}$, and $M_{e}$ denote the hidden representation of phoneme-lip feature, prosody variances, and emotional tone, respectively. %
Specifically, our mel-spectrogram decoder consists of a stack of self-attention layers~\cite{vaswani2017attention} and 1D-convolution layers as in FastSpeech~\cite{ren2019fastspeech}. 
We then use the mel-linear layer and  postnet~\cite{shen2018natural} to refine the hidden states into final dimensional mel-spectrograms by:
\begin{equation}
    \begin{aligned}
        y = \mathrm{PostNet}(\mathrm{FC}(f)).
     \end{aligned}
\end{equation}
Finally, to generate the time-domain waveform $y_w$ from mel-spectrogram $y$, we use HiFi-GAN~\cite{kong2020hifi} as our vocoder, which mainly consists of a transposed convolution network and a multi-receptive field fusion module.

\subsection{Loss Functions}
Our model is trained in an end-to-end fashion via optimizing the sum of all losses:
\begin{equation}
\mathcal{L}_{pitch} = \frac{1}{T_y} \sum \limits_{t=1}^{T_y-1} ({{P}_t}-{\hat{P}_{val}^t})^2,
\end{equation}
\begin{equation}
\mathcal{L}_{energy} = \frac{1}{T_y} \sum \limits_{t=1}^{T_y-1} ({E}_t-{\hat{E}_{aro}^t})^2,
\end{equation}
\begin{equation}
\mathcal{L}_{emo} =  -\sum \limits_{i=1}^C {G_i}\mathrm{log}(\hat{G}_{emo}^i),
\end{equation}
\begin{equation}
\mathcal{L}_{mel} = \frac{1}{T_y} \sum \limits_{t=1}^{T_y-1} \left \| {\mathbf{f}}_t-\hat{\mathbf{f}}_t \right \|.
\end{equation}
\begin{equation}
    \begin{aligned}
        \mathcal{L}_{S} = \lambda_1\mathcal{L}_{mel} + \lambda_2 \mathcal{L}_{pitch} + \lambda_3 \mathcal{L}_{energy} + \lambda_4 \mathcal{L}_{emo},
     \end{aligned}
     \label{eq:synthesizer_loss}
\end{equation}
where $\mathcal{L}_{mel}$,  $\mathcal{L}_{pitch}$, $\mathcal{L}_{energy}$ and $\mathcal{L}_{emo}$, denote the losses of mel-spectrogram, pitch, energy and global emotion respectively. 
${P}_t$, ${E}_t$, and $\mathbf{f}_t$ are ground-truth pitch, energy, and mel-spectrogram on frame-level, respectively. G is the ground-truth emotional label and C denotes all categories. 

\begin{table*}[t]
  \centering
  \resizebox{1.0\linewidth}{!}
  {
    \begin{tabular}{cccccccccc}
    \toprule
    Methods & LSE-D $\downarrow$ & LSE-C $\uparrow$ & MCD $\downarrow$ & MCD-DTW $\downarrow$ & MCD-DTW-SL $\downarrow$ &Id. Acc. $\uparrow$ & Emo. Acc. $\uparrow$ & MOS-N $\uparrow$ & MOS-S $\uparrow$ \\
    \midrule
    Ground Truth  &  6.734  &  7.813  & 00.00 & 00.00  & 00.00 & 90.62  & 84.38  & 4.61 $\pm$ 0.15  & 4.74 $\pm$ 0.12 \\
    \midrule 
    SV2TTS~\cite{jia2018transfer} &   13.733    &   2.725  &  21.08    &   12.87    & 49.56 & 33.62 & 37.19 &   2.03 $\pm$ 0.22 & 1.92 $\pm$ 0.15\\
    SV2TTS*~\cite{jia2018transfer} &   12.617    &   3.349  &  19.38    &   12.73    & 34.51 & 35.18 & 42.05 &   2.07 $\pm$ 0.07 & 2.15 $\pm$ 0.09\\
    Tacotron*~\cite{wang2017tacotron} &   13.475    &   2.938  &  19.79    &   18.73    & 42.15 & 32.49 & 39.68 &   2.12 $\pm$ 0.17 & 2.06 $\pm$ 0.12\\
    FastSpeech2~\cite{ren2020fastspeech} &   12.261   &  2.958   & 20.78 & 14.39 & 19.41 & 21.72  & 46.82 &  2.79 $\pm$ 0.10  & 2.63 $\pm$ 0.09  \\
    FastSpeech2*~\cite{ren2020fastspeech} &   12.113   &  2.604   & 20.66 & 14.59 & 20.79 & 23.44  & 46.90 &  3.08 $\pm$ 0.06  & 2.89 $\pm$ 0.07  \\
    V2C-Net\tablefootnote{V2C republish official results on \href{https://github.com/chenqi008/V2C}{https://github.com/chenqi008/V2C}.}~\cite{chen2022v2c} &   11.784   &  3.026   & 20.61 & 14.23 &19.15 & 26.84  & 48.41 &3.19 $\pm$ 0.04  &3.06 $\pm$ 0.06 \\
    \midrule
     Ours &  \textbf{8.036}    &  \textbf{5.608}   &  \textbf{15.66}  &  \textbf{12.29}  &  \textbf{13.48}  &  \textbf{37.75}  &  \textbf{61.46}  &  \textbf{4.03 $\pm$ 0.08}  & \textbf{3.89 $\pm$ 0.07}  \\
    \bottomrule
    \end{tabular}%
    }
    \vspace{-5pt}
  \caption{Results on the V2C dataset with comparisons against state-of-the-art methods. 
  We provide the results using both objective metrics (\ie, LSE-D, LSE-C, MCD, MCD-DTW and MCD-DTW-SL) and 
   subjective metrics (\ie, MOS-Naturalness and MOS-Similarity).
  ``Id. Acc.'' and ``Emo. Acc.'' are the identity and emotion accuracy of the generated speech, respectively. 
  The method with ``*'' refers to a variant taking video (emotion) embedding as an additional input as in~\cite{chen2022v2c}. $\uparrow(\downarrow)$ means that the higher (lower) value is better.}
  \vspace{-10pt}
  \label{tab1}%
\end{table*}%

\section{Experimental Results}
In this section, we first briefly describe the datasets used for evaluation and the evaluation metric. Then we present the implementation details of the proposed method. Last, we show the results compared to state-of-the-art methods and the ablation study.

\subsection{Datasets}
\smallskip \noindent\textbf{V2C} is a multi-speaker dataset for animation movie dubbing with identity and emotion annotations~\cite{chen2022v2c}. 
It is collected from 26 Disney cartoon movies and covers 153 diverse characters. 
V2C not only needs to generate voices with identity characteristics according to the reference audio but also capture emotional information based on the reference movie clips. 
The whole dataset has 10,217 video clips with paired audio and subtitles. %
The training/validation/test size are 60\%, 10\%, 30\%.

\noindent\textbf{Chem} is a single-speaker dataset composed of 6,640 short video clips collected from the YouTube, with the total video length of approximately nine hours~\cite{hu2021neural}. 
The Chem dataset is originally used for the unconstrained single-speaker lip-to-speech synthesis~\cite{prajwal2020learning}, which takes place in a chemistry lecture. 
For fluency and complete dubbing, each video clip has sentence-level text and audio based on the start and end timestamps. %
There are 6,240, 200, and 200 dubbing clips for training, validation, and testing, respectively.

\subsection{Evaluation Metrics}
\noindent\textbf{Audio-visual synchronization.} 
To evaluate the synchronization between the generated speech and the video quantitatively, we adopt Lip Sync Error Distance (LSE-D) and Lip Sync Error Confidence (LSE-C) as our metrics, which can explicitly test for synchronization between lip motions and speech in unconstrained videos in the wild~\cite{prajwal2020lip,chung2016out}.

\noindent\textbf{Mel Cepstral Distortion and its variants.} 
MCD~\cite{kubichek1993mel}, MCD-DTW~\cite{battenberg2020location} and MCD-DTW-SL~\cite{chen2022v2c} are adopted, which reflect the similarity of mel-spectrograms. 
MCD-DTW uses the Dynamic Time Warping (DTW)~\cite{muller2007dynamic} algorithm to find the minimum MCD between two speeches, while MCD-DTW-SL introduces the duration measure coefficient to consider the length and the quality of generated speech~\cite{chen2022v2c}. 

\noindent\textbf{Emotion and identity accuracy.}
To measure whether the generated speech carries proper emotion and speaker identity, we adopt an emotion accuracy (Emo. Acc.) and an identity accuracy (Id. Acc.) as our metrics as in~\cite{chen2022v2c}.

\noindent\textbf{Subjective evaluations.} To further evaluate the quality of generated speech, we conduct a human study using a subjective evaluation metric, following the settings in~\cite{chen2022v2c}. 
Specifically, we adopt the MOS-naturalness (MOS-N) and MOS-similarity (MOS-S) to assess the naturalness of the generated speech and the recognization of the desired voice.

\subsection{Implementation Details}
For the duration aligner, we use 4 Feed-Forward Transformer (FFT) blocks and 3 FFT blocks for the phoneme encoder and lip movement encoder, respectively. 
We set the dimension of the phoneme feature $\mathcal{O}$ and lip features $E_{lip}$ to 256. %
We use 8 attention heads for alignment between lip and phoneme.
For each movie clip, Our $FPS$ set is 25, the sampling rate $sr$ is 22050$Hz$. We use short-time fourier transform (STFT) to obtain the mel-spectrum, and the number of points of the fourier transform is 1024. %
We use the Conv-Transpose1D module with
2 stride and 4 kernel sizes to obtain the duration features. %
In Valence and Arousal Feature Encoder, the EmoFAN consists of one 2D convolution with a kernel size of 7 × 7  and 3 convolution blocks (ConvBlock) with a kernel size of 3 × 3 and Average Pooling stride of 2 × 2.
Similarly, we set the dimension of the valence and arousal feature to 256.
For the mel-spectrograms generator, the mel-spectrogram decoder consists of 6 FFT blocks and the hidden state of mel-linear layer is of size 80.

For training, we use Adam~\cite{kingma2014adam} with $\beta_1$ = 0.9, $\beta_2$ = 0.98, $\epsilon$=$10^{-9}$ to optimize our model. %
For V2C and Chem dataset, we set the learning rate schedule to 0.00001 and 0.00005, respectively.
In this work, we use pretrained HiFiGAN~\cite{kong2020hifi} as the vocoder to transform the generated mel-spectrograms into audio samples. %
We set the batch size to 16 on two datasets. %
Our model is implemented in PyTorch~\cite{paszke2019pytorch}. All the models are performed on a single NVIDIA GTX3090Ti GPU. %
We train the model with 600 epochs on the V2C dataset and 400 epochs on the Chem dataset.

\subsection{Quantitative Evaluation}
We compare with five related baselines of speech synthesizes.
(1) SV2TTS~\cite{jia2018transfer} is a basic TTS model to generate speech with reference audio for multi-speakers; (2) Tacotron~\cite{wang2017tacotron} is an end-to-end generative TTS model that synthesizes speech directly from textual characters;   (3) FastSpeech2~\cite{ren2020fastspeech} introduces the variance adaptor to convert the text to waveform by end-to-end; (4) Neural Dubber~\cite{hu2021neural} synthesizes human speech for given video according to the corresponding text; (5) V2C-Net~\cite{chen2022v2c} is the first model for movie dubbing, which match the speaker’s emotion presented in video. 
Note that we do not compare with Neural Dubber~\cite{hu2021neural} on the V2C benchmark due to the unavailability of its code and missing implementation details.

\noindent\textbf{Results on the V2C benchmark.} 
The results are presented in Table~\ref{tab1}. Our method achieves the best performance  on all nine metrics. 
Specifically, in terms of audio-visual sync, our method achieves 8.036 of LSE-D and 5.608 of LSE-C, which significantly surpasses the previous best results and is much closer to human performance. 
In terms of MCD, MCD-DTW, and MCD-DTW-SL, our method achieves relative 24.02\%, 13.63\% and 29.61\% improvements, respectively. 
This indicates our method can achieve a better mel-spectrogram than others. 
The above results together show that bridging specific attributes of speech with corresponding visual counterparts can make the generated speech present better prosodies and   lip motion sync. 
Additionally, our method outperforms the previous method by a large margin in emotion accuracy and can gain better identification accuracy.  
This indicates the proposed method can better capture and convey emotions, which is of great importance for the movie dubbing task.
Last, the human subjective evaluation results (see MOS-N and MOS-S) also show that our method can generate speeches that are closer to realistic speech according to naturalness and similarity. 

\begin{table}[!t]
  \centering
  \resizebox{1.0\linewidth}{!}
  {
  \begin{tabular}{ccccc}
    \toprule
        Methods & AQ $\uparrow$ & AV Sync $\uparrow$ & LSE-D $\downarrow$ & LSE-C $\uparrow$ \\
    \midrule
        Ground Truth & 3.93 $\pm$  0.08 & 4.13$\pm$0.07 & 6.926  & 7.711\\
    \midrule
    	FastSpeech2~\cite{ren2020fastspeech} &3.71$\pm$0.08 &3.29$\pm$0.09 & 11.86 & 2.805 \\
    	Tacotron*~\cite{wang2017tacotron} &3.55$\pm$0.09 &3.03$\pm$0.10 & 11.79 & 2.231 \\
    	V2C-Net~\cite{chen2022v2c} & 3.48$\pm$0.14 & 3.25$\pm$0.11 &  11.26 & 2.907 \\
        Neural Dubber~\cite{hu2021neural} &3.74$\pm$0.08 &3.91$\pm$0.07 & 7.212 & 7.037 \\
    \midrule
    Ours &  \textbf{3.84$\pm$0.11}    &  \textbf{3.97$\pm$0.08}   &  \textbf{6.975}  &  \textbf{7.176} \\
    \bottomrule
  \end{tabular}
  }
  \caption{Results on the Chem dataset with comparisons against state-of-the-art methods. AQ (Audio Quality) and AV Sync (audio-visual synchronization) are subjective metrics. Note that the Chem dataset is a single-speaker non-movie dataset, and thus there is no identity accuracy and emotion accuracy.
  }
  \label{tab:chem}
\end{table}

\noindent\textbf{Results on the Chem benchmark.} 
As shown in Table~\ref{tab:chem}, our model is ahead of the state-of-the-art methods in all metrics on the Chem benchmark.
In terms of the audio-visual sync, our method achieves 6.975 LSE-D and 7.176 LSE-C. 
Furthermore, in the subjective evaluations, our method improves 10.34\% on AQ and 22.15\% on AV Sync. The results show that our performance is much closer to the ground truth recording, which indicates that our model synthesizes high-quality natural speech by controlling the prosody from hierarchical visual representation.

\begin{table}[!tbp]
  \centering
  \resizebox{1.0\linewidth}{!}
  {
    \begin{tabular}{lcccccc}
    \toprule
    \# & Methods  & LSE-D $\downarrow$ & LSE-C $\uparrow$ & MCD $\downarrow$ &Id. Acc. $\uparrow$ & Emo. Acc. $\uparrow$ \\
    \midrule
    1 & w/o DA &  11.835  &   3.716 & 19.53 & 18.75  & 38.33  \\
    2 & w/o PA &  8.514  &   5.274 & 16.34  & 10.42  & 22.08  \\
    3 & w/o AB &  8.261  &   5.408 & 15.90  & 33.33  & 53.92 \\
    \midrule
    4 & w/o Valence &  9.215  &  4.935 & 16.31  & 29.81  & 35.31 \\
    5 & w/o Arousal &  8.793  &   5.216 & 15.79  & 32.74  & 46.38 \\
    \midrule
    6 & VA $v.s.$ FF &  9.160  &   5.011 & 19.87  & 23.67  & 39.58 \\
    \midrule
    7 & w/o multi-head &  10.948  &   3.894 & 18.65  & 24.85  & 40.25 \\
    8 & Duplication &  11.475  &   3.814 & 18.63  & 21.78  & 37.92 \\
    
    \midrule
    9 & Full model  & \textbf{8.036}    &  \textbf{5.608}   &  \textbf{15.66} & \textbf{37.75}  &  \textbf{61.46}    \\ 
    \bottomrule
    \end{tabular}%
    } 
  \caption{Ablation study of the proposed method on the V2C benchmark dataset.} 
  \vspace{-10pt}
  \label{tab_ablation}%
\end{table}%

\subsection{Ablation Studies}

\noindent\textbf{Effectiveness of Duration Aligner, Prosody Adaptor, and Atmosphere Booster}. 
We evaluate the effectiveness of these three modules by removing them separately and retraining the model. 
The results are shown in Row 1$\sim$3 of Table~\ref{tab_ablation}. 
The result shows that all the proposed modules contribute significantly to the overall performance, and each module has a different focus. 
Specifically, the performance on the audio-visual metrics (LSE-D, LSE-C, and MCD) drops the most when removing the Duration Aligner (DA). This reflects the DA module indeed helps the model learn a better temporal synchronization. 
By contrast, the performance on identity accuracy and emotion accuracy drop the most when removing the Prosody Adaptor (PA). This can be attributed to the predicted pitch and energy representing a speaker's identity and his/her emotion. 
When removing the Atmosphere Booster (AB), the performance drops compared to the full model (Row 9) but does not drop as much as when removing the other two modules. This indicates the AB module also contributes the overall performance improvement but contributes the least in the three modules.

\vspace{1mm}
\noindent\textbf{Effectiveness of valence and arousal}.
In our model, we bridge arousal with energy and valence with pitch by attention mechanism. To evaluate their effectiveness, we cut off the connection and predict energy and pitch directly from the phoneme-lip representation. 
The results are presented in Row 4$\sim$5 of Table~\ref{tab_ablation}.
It shows that the  performance drops significantly when removing either of them and drops more when removing valence. 
This indicates valence contributes more than arousal on the V2C task. 

\begin{figure}[!t]
    \centering
    \includegraphics[width=1.0\linewidth]{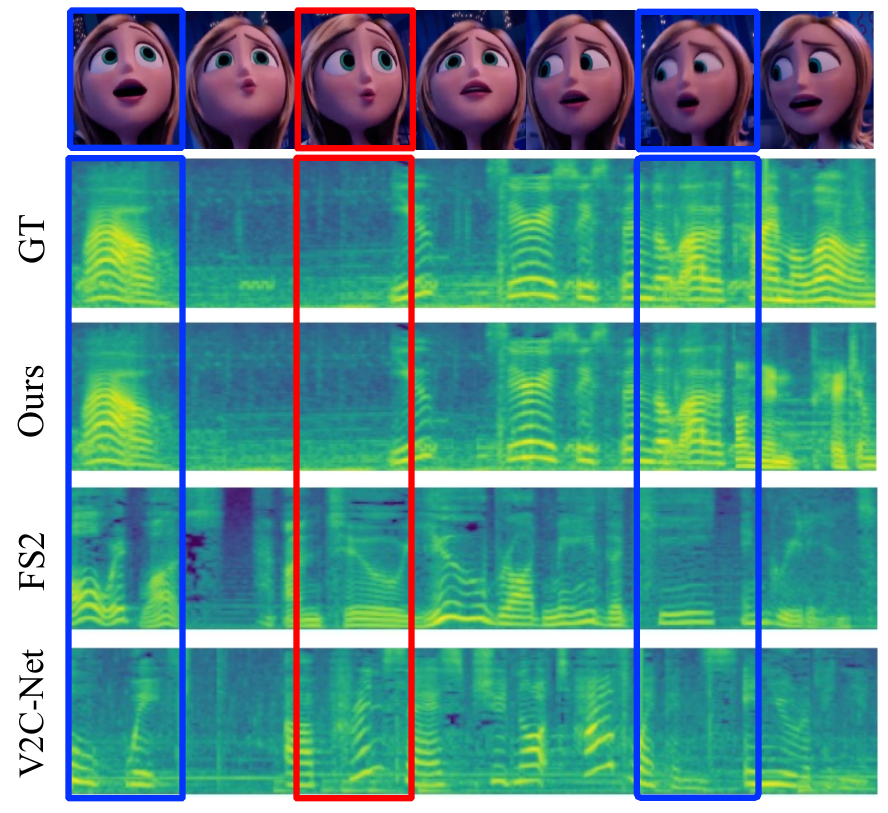}
    \caption{ Audiovisual consistency visualization on V2C dataset: Ground Truth (GT), our model, FastSpeech 2 (FS2) and V2C-Net.
    }
    \label{fig_3}
    \vspace{-10pt}
\end{figure}

\begin{figure*}[t]
    \centering
    \includegraphics[width=1.0\linewidth]{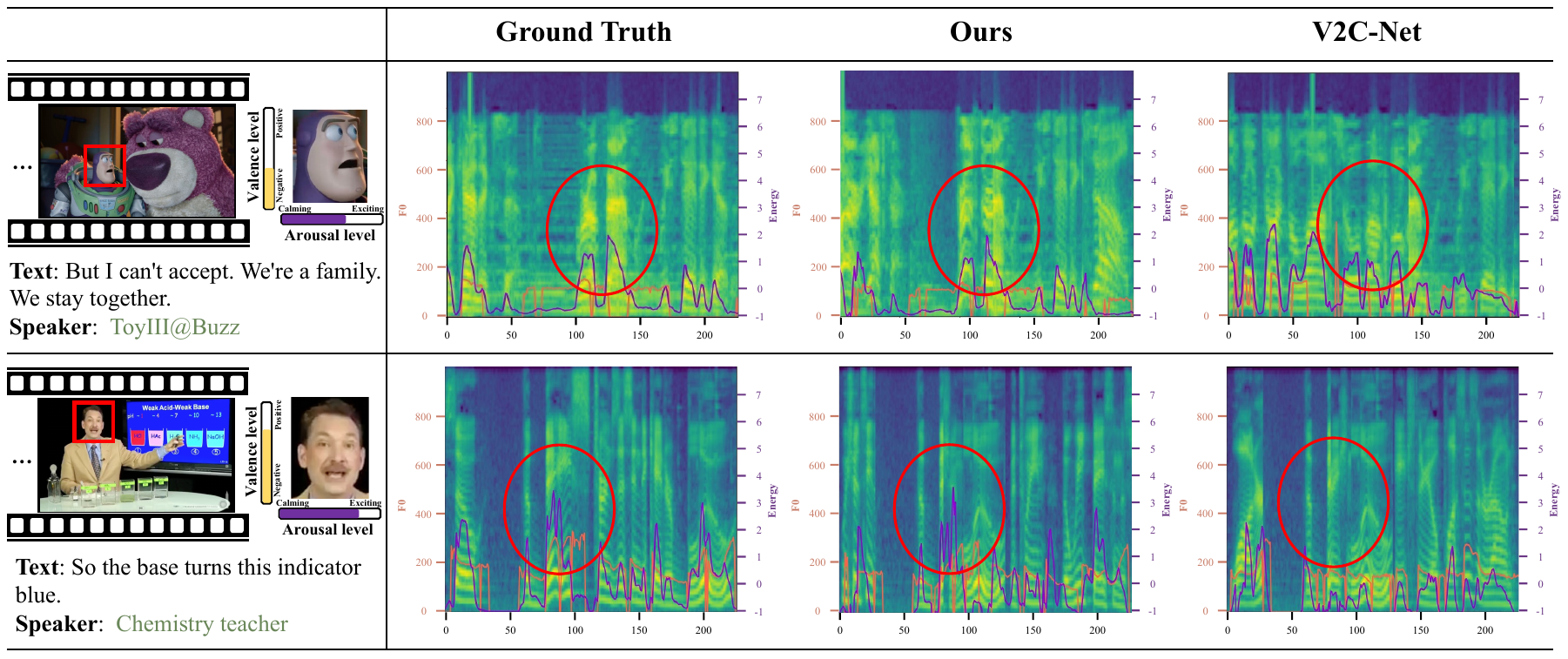}
    \caption{Visualization of audio  on V2C dataset (top) and Chem dataset (bottom). Orange curves are $F_0$ contours, where  $F_0$ is the fundamental frequency of audio. Purple curves refer to the energy (volume) of audio. The horizontal axis is the duration of the audio. The red circles highlight the mel-spectrograms at the same moment as the frame shown on the left side.
    }
    \vspace{-10pt}
    \label{fig4}
\end{figure*}
\vspace{1mm}
\noindent\textbf{Valence-and-Arousal v.s. Facial Expression}.
To compare the role of affective display and facial expressions on prosody inference, we replace the input of the APA module with original embeddings of facial features from CNN. 
As row “VA $v.s.$ FF” of Table~\ref{tab_ablation} shows, the model performance drops significantly. %
This is likely caused by facial features that are still far from the information-containing emotion, which is not enough to guide prosody generation. 

\vspace{1mm}
\noindent\textbf{Multi-head Attention in Duration Adaptor}. In our duration aligner, we exploit multi-head attention to learn the relation between the phoneme sequences and the lip motion.
To verify its effectiveness, we conduct experiments using conventional dot-product attention.
The results are presented in Row 7 of Table~\ref{tab_ablation}. 
The performance drops significantly on all metrics, such as 26.59\% and 44.02\% decrease on LSE-D and LSE-C compared to Row 9, respectively. This indicates multi-head attention learns a much better correlation between phonemes with lip movement.

\vspace{1mm}
\noindent\textbf{Conv-Transpose v.s. Duplication}.
In our duration aligner, we propose to use Conv-Transpose to learn the expansion of the fused phoneme-lip representation to its desired length. 
To verify its effectiveness, we replace it with simple make $n$ duplications as in~\cite{hu2021neural}. The results are shown in Row 8 of Table~\ref{tab_ablation}. It shows that the performance drops significantly. For example, it falls {15.94\% on MCD metric}. This demonstrates the superiority of using transposed convolution to learn the upsampling than simply copying.

\subsection{Qualitative Results}
\noindent\textbf{Audiovisual consistency visualization}.
Figure~\ref{fig_3} presents the mel-spectrograms of generated audios along with its frames.
Blue and red bounding boxes denote whether the character is speaking or not, respectively.
Compared with other methods, the mel-spectrogram generated by our model is closer to the ground truth, indicating better audio-visual synchronization. %
This can be attributed to our duration aligner, which exploits multi-head alignment between the text phoneme sequence and the lip movement sequence.
By controlling the lip movements explicitly, we obtain the desired length of mel-spectrograms, which makes the speech well synchronized with the input video.

\vspace{1mm}
\noindent\textbf{Arousal and valence with prosody visualization}.
We selected two examples from the test set of the V2C dataset and Chem dataset to demonstrate the alignment between energy and arousal as well as pitch and valence.
The valence (positive or negative) and arousal  (calming or exciting) of facial expressions are shown in the first column. 
The main pitch and energy are shown in orange and blue curves in the right column.
We use the red circle to highlight the pitch and energy that correspond to the video frame shown in the left column.
The result shows that our method achieves energy and pitch closer to the ground truth speech. 
When the chemistry teacher becomes excited and positive, our model successfully leverages the affective display to synthesize a similar pitch and energy as the ground-truth speech.

\section{Conclusion}
In this work, we propose a hierarchical prosody modeling network for movie dubbing, which bridges video representations and speech attributes from three levels: lip, facial expression, and scene. %
By associating these visual representations with their voice counterparts, we obtain more powerful representations for dubbing. 
Furthermore, we design an affective-display based prosody adaptor, which effectively learns to align the valence and arousal to  the pitch and energy of speeches.
Our proposed model sets new state-of-the-art on both Chem and V2C-Animation benchmarks.

\textbf{Acknowledgement.} This work was supported in part by the National Key R\&D Program of China under Grant 2018AAA0102000, and in part by National Natural Science Foundation of China under Grants 62225207 and U19B2038, the Key R\&D Plan Project of Zhejiang Province (No. 2023C01004),  Shandong Provincial
Natural Science Foundation (ZR2020MA064), Youth Innovation Promotion Association of CAS under Grant 2020108.

{\small
\bibliographystyle{ieee_fullname}
\bibliography{ref}
}

\end{document}